\documentclass{article}
\usepackage{ijcai17}
\usepackage{graphicx}
\usepackage[all, knot]{xy}
\usepackage{comment}
\usepackage{amssymb,amsmath,latexsym}
\usepackage{times}
 \usepackage[usenames]{color} 
\newtheorem{example}{Example}
\newtheorem{definition}{Definition}

\newtheorem{theorem}{Theorem}
\newtheorem{lem}{Lemma}
\newtheorem{proof}{Proof}

\newcommand{\tuple}[1]{\langle #1 \rangle}  
\newcommand{\att}{\rightarrow}  
  
\newcommand{\AR}{\mathcal{A}} 
\newcommand{\attack}[2]{#1\rightarrow #2}  
  
\newcommand{\attackees}[1]{#1^\rightarrow} 
\newcommand{\attackers}[1]{#1^\leftarrow} 
\newcommand{\dg}[1]{\mathrm{d}(#1)} 
\newcommand{\dgn}[1]{ \mathrm{d}(#1)} 
\newcommand{\defender}[1]{\mathbf{defender}(#1)}  
\newcommand{\defendee}[1]{\mathbf{defendee}(#1)}  
\newcommand{\attacker}[1]{\mathbf{attacker}(#1)} 

\title{Defense semantics of argumentation: revisit}
\author{Beishui Liao  \\ 
 Zhejiang University\\
 \And 
 Leendert van der Torre\\ University of Luxembourg\\ }

\begin{document}

\maketitle

\begin{abstract}
 
In this paper we introduce a novel semantics,  called defense semantics,  for Dung's abstract argumentation frameworks in terms of a notion of (partial) defence,  which is a triple encoding that one argument is (partially) defended by another argument via attacking the attacker of the first argument.  
In terms of defense semantics,   we show that defenses related to self-attacked arguments and arguments in 3-cycles are unsatifiable under any situation and therefore can be removed without affecting the defense semantics of an AF.  
Then,  we introduce a new notion of defense equivalence of AFs,  and compare defense equivalence with standard equivalence and strong equivalence, respectively.   Finally,  by exploiting defense semantics,  we define two kinds of reasons for accepting arguments, i.e., direct reasons and root reasons,  and a notion of root equivalence of AFs that can be used in argumentation summarization. 

\end{abstract}

\section{Introduction}
In the field of AI,  abstract argumentation is mainly about the acceptability of arguments in an argumentation framework (AF) \cite{DBLP:journals/ai/Dung95,Baroni:KER,DBLP:journals/ai/CharwatDGWW15}.   
A set of arguments that is collectively acceptable according to some criteria is called an extension.   There are two basic criteria for defining all kinds of extensions that are based on the notion of admissible set,  called conflict-freeness and defense.  An argument is defended by a set of arguments,  if every attacker of this argument is attacked by at least one argument in this set.  Obviously,  the notion of defense plays an important role in evaluating the status of arguments.  However,  this usage of the classical notion of defense has not fully reflected some useful information implicitly encoded by the interaction relation between arguments.  The later can be used to deal with some important problems in formal argumentation.  Let us consider two intereating examples. 

The first example is about the treatment of odd cyles \cite{DBLP:journals/ai/BaumannBU22}.   It has been shown that the computation time is highly related to the existence of cyles,  especially  odd cycles \cite{DBLP:journals/amai/Liao13}.   So,  a natural question arises: whether it is possible to improve the efficiency of computation by exploiting the local information encoded by odd cycles? 

In $\mathcal{F}_{1}$,  the acceptance of $d$ and $c$ can be defined in terms of Dung's argumentation semantics,  but it can be also defined accoding to the following notion of (partial)  defense.   We say that argument $x$ (partially) defends argument $z$ with respect to argument $y$,  if $x$ attacks $y$ and $y$ attacks $z$,  denoted as $(x, y, z)$ or $z_\mathbf{y}^x$.   
For argument $c$,  there are two defenses $c_\mathbf{b}^d$ and $c_\mathbf{b}^a$.  Intuitively,  the defense $c_\mathbf{b}^a$ does not contribute to the acceptance of $c$,  because the self-attacked argument $a$ cannot be accepted in any situation and therefore cannot provide support to the acceptance of some other arguments.  In other words,  if one uses defenses to evaluate the status of arguments,  he may determine the status of $c$ only according to defense $c_\mathbf{b}^d$,  without considering defense $c_\mathbf{b}^a$,  i.e.,  $c_\mathbf{b}^a$ can be removed from the set of defenses.   Here,  one may argue that the self-attacked argument $a$ can be removed,  which also does not affact the staus of $c$.  This is true,  but it is not the case that a self-attacked argument can be removed in any situation. 

  
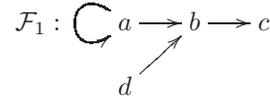
\begin{picture}(206,40)
\put(0,30){\xymatrix@C=0.55cm@R=0.38cm{
\mathcal{F}_{1}: &a\ar@(ul,dl)[]\ar[r]&b\ar[r]&c
\\
 &d\ar[ur]
 }}
\end{picture}

Let's consider the following AF.   If we remove the self-attacked argument $a$,  it is obvious that some other arguments will be affacted.  However,  in the set of defenses $\{a_\mathbf{a}^a$,    $b_\mathbf{a}^a$,    $c_\mathbf{b}^a$,    $d_\mathbf{c}^b\}$,   one may remove some of  defenses in this set without affecting the evaluation of the status of the remaining defenses.  In this example,  one may remove  $a_\mathbf{a}^a$,    $b_\mathbf{a}^a$ and $c_\mathbf{b}^a$,  obtaining a subset of defenses $\{d_\mathbf{c}^b\}$.  By properly define the conditions of accepting a defense,   the set of accepatble defenses of $\{d_\mathbf{c}^b\}$ is emptyset,  which is equivalent to the one when consider the whole set of defenses.  Please refer to Section \ref{ref-section-ucd} for details. 

\begin{picture}(206,25)
\put(0,10){\xymatrix@C=0.55cm@R=0.38cm{
\mathcal{F}_{2}: &a\ar@(ul,dl)[]\ar[r]&b\ar[r]&c\ar[r]&d
 }}
\end{picture}

The second example is about equivalence between AFs.   
In terms of extension-based semantics,  AFs $\mathcal{F}_3$ and $\mathcal{F}_4$ are obviously not equivalent.  However, if we consider the indirect reasons of acceptance of a given set of arguments based on the notion of defense,  they are equivalent.  More specifically,  in $\mathcal{F}_3$, accepting $a$ is a reason to accept $d$,  because $a$ defends $d$.  Similarly,
  accepting $d$ is a reason to accept $e$, and  accepting $e$ is a reason to accept $a$. If we allow this relation to be transitive, we find that accepting $a$ is a reason to accept $a$. Similarly,  accepting $b$ is a reason to accept $b$. Meanwhile, in $\mathcal{F}_4$,  we have: accepting $a$ is a reason to accept $a$, and accepting $b$ is a reason to accept $b$.  So, from the perspective of the reasons for accepting $a$ and $b$,  $\mathcal{F}_4$ is equivalent to $\mathcal{F}_3$, or $\mathcal{F}_4$  is a summarization of  $\mathcal{F}_3$ from another point of view. 

 
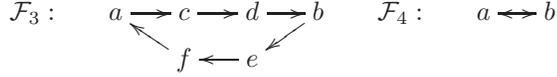
\begin{picture}(206,35)
 \put(0,25){\xymatrix@C=0.5cm@R=0.1cm{
  \mathcal{F}_3: &a\ar[r]&c\ar[r]&d\ar[r]&b\ar[ld]& \mathcal{F}_4: & a\ar[r]&b\ar@/^0.0cm/[l] \\
 & &f\ar[lu] &e\ar[l] &  }}  
    
    \end{picture}


Now,  consider the question when two AFs are equivalent in a dynamic setting.  For $\mathcal{F}_5$ and $\mathcal{F}_6$ below,  both of them have a complete extension $\{a, c\}$. However, the reasons of accepting $c$ in $\mathcal{F}_5$ and $\mathcal{F}_6$ are different. For the former, $c$ is defended by $a$, while for the latter,  $c$ is unattacked and has no defender. In this sense, $\mathcal{F}_5$ and $\mathcal{F}_6$ are not equivalent. For example, in order to change the status of argument $c$ from ``accepted'' to ``rejected'', in $\mathcal{F}_5$, one may produce a new argument to attack the defender $a$,  or to directly attack $c$. However, in $\mathcal{F}_6$ using an argument to attack $a$ cannot change the status of $c$, since $a$ is not a defender of $c$.

  \begin{picture}(206,24)
 \put(0,10){\xymatrix@C=0.6cm@R=0.2cm{
  \mathcal{F}_5: \hspace{0.3cm} a\ar[r]&b\ar[r]&c &\hspace{0.5cm} \mathcal{F}_6:  \hspace{0.3cm}  a\ar[r]&b&c  
   }}
    
 \end{picture}

In terms of the above analysis,  one question arises: under what conditions, can two AFs be viewed as equivalent? The existing notions of argumentation equivalence, including standard equivalence and strong equivalence, are not sufficient to capture the equivalence of the AFs in the situations mentioned above.  More specifically, $\mathcal{F}_3$ and $\mathcal{F}_4$ are not equivalent in terms of the notion of standard equivalence or that of strong equivalence,  but they are equivalent in the sense that the reasons for accepting arguments $a$ and $b$ in these two graphs are the same. $\mathcal{F}_5$ and $\mathcal{F}_6$ are equivalent in terms of standard equivalence, but they are not equivalent in the sense that the reasons for accepting $c$ in these two graphs are different.  Although the notion of strong equivalence can be used to identify the difference between $\mathcal{F}_5$ and $\mathcal{F}_6$, conceptually it is not defined from the perspective of reasons for accepting arguments,  while the latter is formualted in terms of defenses between arguments.

Motivated by the above intuitions,  we propose a novel semantics of argumentaion in terms of a new notion of defense.

The structure of this paper is as follows. In Section 2, we introduce some basic notions of argumentation semantics. In Section 3, we propose a new notion of defense.  In Section 4, we formulate defense semantics. In Section 5,  we introduce unsatisfiabilty and contraction of defenses.   In section 6,
we introduce new equivanlence raltions between AFs.  Finally,  we conclude in Section 7.

\section{Dung's semantics}
An AF is defined as $\mathcal{F} = (\AR, \att)$, where $\AR$ is a set of arguments and $\att\subseteq \AR\times \AR$ is a set of attacks between arguments. 

Let $\mathcal{F} = (\AR, \att)$ be an AF.  Given a set $B\subseteq \AR$ and an argument $\alpha\in \AR$, $B$ attacks $\alpha$, denoted $\attack{B}{\alpha}$,  iff there exists $\beta\in B$ such that $\attack{\beta}{\alpha}$. 
Given an argument $\alpha\in \AR$, let  $\attackers{\alpha} = \{\beta\in \AR \mid \attack{\beta}{\alpha}\}$ be the set of arguments attacking $\alpha$, and $ \attackees{\alpha} = \{\beta\in \AR \mid \attack{\alpha}{\beta}\}$ be the set of arguments attacked by $\alpha$. When $\attackers{\alpha} = \emptyset$, we say that $\alpha$ is unattacked, or $\alpha$ is an initial argument. 

Given $\mathcal{F} = (\AR, \att)$ and $E\subseteq \AR$, we say: $E$  is \emph{conflict-free} if $\nexists \alpha, \beta\in E$ such that $\attack{\alpha}{\beta}$; $\alpha\in \AR$ is \textit{defended} by $E$  if $\forall\attack{\beta}{\alpha}$, it holds that $\attack{E}{\beta}$; $B$ is \textit{{admissible} } if  $E$ is  conflict-free, and each argument in $E$ is defended by $E$; $E$ is a \textit{complete extension} of $\mathcal{F}$ if $E$ is admissible, and each argument in $\AR$ that is defended by $E$ is in $E$.   $E$ is a \textit{preferred extension} of $\mathcal{F}$ if $E$ is an maximal complete extension of $\mathcal{F}$.    $E$ is the \textit{grounded extension} of $\mathcal{F}$ if $E$ is the minimal complete extension of $\mathcal{F}$ .  
W use $\sigma(\mathcal{F})$  to denote the set of $\sigma$ extensions of $\mathcal{F}$,  where $\sigma\in\{\mathrm{co}, \mathrm{pr},\mathrm{gr},\mathrm{st}\}$ is a function mapping each AF to a set of $\sigma$ extensions, called $\sigma$ semantics.  

For AFs $\mathcal{F}_1= (\AR_1, \rightarrow_1)$  and $\mathcal{F}_2 = (\AR_2, \rightarrow_2)$, we use $\mathcal{F}_1\cup \mathcal{F}_2$ to denote $ (\AR_1\cup \AR_2, \rightarrow_1\cup \rightarrow_2)$.  The standard equivalence and strong equivalence of AFs are defined as follows. 

\begin{definition}[Standard and strong equivalence of AFs]  \cite{DBLP:conf/kr/OikarinenW10} \label{def-sequi}
Let  $\mathcal{F}$  and $\mathcal{G}$ be two AFs.
\begin{itemize}
\item $\mathcal{F}$  and $\mathcal{G}$ are of standard equivalence w.r.t. a semantics $\sigma$, in symbols $\mathcal{F} \equiv^\sigma \mathcal{G}$, iff $\sigma(\mathcal{F}) = \sigma(\mathcal{G})$.
\item $\mathcal{F}$  and $\mathcal{G}$ are of strong equivalence w.r.t. a semantics $\sigma$, in symbols $\mathcal{F} \equiv_s^\sigma \mathcal{G}$, iff for all AF $\mathcal{H}$, it holds that $\sigma(\mathcal{F}\cup \mathcal{H}) = \sigma(\mathcal{G}\cup \mathcal{H})$. 
\end{itemize}
\end{definition} 

\begin{example}
Consider $\mathcal{F}_1 - \mathcal{F}_4$ in Section 1.  In terms of Definition \ref{def-sequi}, under complete semantics, we have: $\mathcal{F}_3\not\equiv^\mathrm{co} \mathcal{F}_4$, $\mathcal{F}_3\not\equiv_s^\mathrm{co} \mathcal{F}_4$; $\mathcal{F}_5\equiv^\mathrm{co} \mathcal{F}_6$, $\mathcal{F}_5\not\equiv_s^\mathrm{co} \mathcal{F}_6$. 
\end{example}

Given an AF $\mathcal{F} = (\AR, \rightarrow)$, the kernel of $\mathcal{F}$ under complete semantics, call \textit{c-kernel},  is defined as follows. 

\begin{definition}[c-kernel of an AG] \cite{DBLP:conf/kr/OikarinenW10}
For an AF $\mathcal{F} = (\AR, \rightarrow)$,  the c-kernel of  $\mathcal{F}$ is defined as $\mathcal{F}^\mathrm{ck} = (\AR, \rightarrow^\mathrm{ck})$,  where
\begin{eqnarray}
\rightarrow^\mathrm{ck}  &=&  \rightarrow \setminus \{\attack{\alpha}{\beta} \mid \alpha \neq \beta, \attack{\alpha}{\alpha}, \attack{\beta}{\beta}\}
\end{eqnarray}
\end{definition}

According to \cite{DBLP:conf/kr/OikarinenW10}, it holds that $\mathrm{co}(\mathcal{F}) = \mathrm{co}(\mathcal{F}^\mathrm{ck})$.

\section{Defense}
In this section, we introduce a notion of defense.  

\begin{definition}[Defense] \label{def-dr}
Let $\mathcal{F} = (\AR, \att)$ be an AF. 
A (partial) defense is a triple $(x, y, z)$ such that $x$ attacks $y$ and $y$ attacks $z$, where $x, y, z\in \AR$. 
We use $z^x_{\mathbf{y}}$ to denote that $x$ is a defender of a defendee $z$ with respect to an attacker $y$.  For an initial argument $z$,  we use $(\top, z)$ (resp., $z^\top$) to denote that $z$ is always defended given that it has no attacker.  For an argument $z$ that is attacked by an initial argument $y$,  we use $(\bot, y,z)$ (resp., $z^\bot_{\mathbf{y}}$) to denote that $z$ is not defended by any argument  with respect to an attacker $y$.  
\end{definition}

The set of defenses of $\mathcal{F}$ is denoted as $\dgn{\mathcal{F}}$.

Given a defense $z^x_{\mathbf{y}}$,  we use the following notatioins to denote the defendee,  defender and attacker in the defense: 
\begin{itemize}
\item $\defendee{z^x_{\mathbf{y}}} = z$
\item $\defender{z^x_{\mathbf{y}}} = x$
\item $\attacker{z^x_{\mathbf{y}}} = y$
\end{itemize}

Given a set of defenses $D\subseteq \dgn{\mathcal{F}}$,  we use the following notations:
\begin{itemize}
\item $\defendee{D} = \{ z \mid z^x_{\mathbf{y}}\in D  \}$
\item $\defender{D} =  \{  x \mid z^x_{\mathbf{y}}\in D \}$
\item $\attacker{D} =  \{  y \mid z^x_{\mathbf{y}}\in D \}$
\end{itemize}

\begin{example}
Consider $\mathcal{F}_1 - \mathcal{F}_4$, we have:
\begin{itemize}
\item $\dgn{\mathcal{F}_1} = \{a^{e}_{\mathbf{f}},  {c}^{f}_{\mathbf{a}},  {d}^{a}_{\mathbf{c}},  {b}^{c}_{\mathbf{d}},  {e}^{d}_{\mathbf{b}}, {f}^{b}_{\mathbf{e}} \}$.
\item  $\dgn{\mathcal{F}_2} = \{a^{a}_{\mathbf{b}}, {b}^{b}_{\mathbf{a}}\}$.
\item  $\dgn{\mathcal{F}_3} = \{a^{\top},  {b}^{\bot}_{\mathbf{a}}, {c}^{a}_{\mathbf{b}}\}$.
\item  $\dgn{\mathcal{F}_4} = \{a^{\top},  {b}^{\bot}_{\mathbf{a}},  {c}^{\top}\}$.
\end{itemize}
\end{example}

\section{Defense semantics}
\begin{definition}[Defense semantics]
Let $U$ be the universe of arguments.  Defense semantics is defined as a partial function $\mathit{def}:  2^{U\times U\times U} \rightarrow 2^{2^{U\times U\times U}}$,  which  associates a set of defenses with a set of subsets of defenses. 
\end{definition}

\begin{definition}[Admissible set of defenses]\label{def-ext-of-defences}
Let $\mathcal{F} = (\AR, \att)$ be an AF.  and $\dgn{\mathcal{F}}$ the  set of defenses of $\mathcal{F}$.  Given $D\subseteq \dgn{\mathcal{F}}$,
$D$ is admissible iff it satisfies the following conditions:
\begin{itemize}
\item[i.] $\defendee{D} \cap \attacker{D} = \emptyset$,
\item[ii.] $\bot\notin \defender{D}$,
\item[iii.] for all $u\in \defender{D}$: if $u\neq \top$ and $u\neq \bot$ then $u\in \defendee{D}$,  
\item[iv.] for all $z^x_{\mathbf{y}}, z^{x^\prime}_{\mathbf{y^\prime}} \in   \dgn{\mathcal{F}}$  if $\mathbf{y}\not= \mathbf{y^\prime}$ and $x^\prime\neq \bot$,  then: if $z^{x}_{\mathbf{y}}\in  D$ and there exists no $z^{x^{\prime\prime}}_{\mathbf{y^\prime}}\in  D$,  then  $z^{x^\prime}_{\mathbf{y^\prime}}\in  D$.
\end{itemize}
\end{definition}

In this definition,  the first item says that no defendee in $D$ is attacked by any attacker in $D$.  The second item means that $\bot$ cannot be used as a defender.  The third item requires that each defender in $D$ should also be a defendee in $D$ if it is not $\top$ or $\bot$.  The fourth item says that for two  defenses with different attackers,   if one is in an admissible set,  then another whose defender is not $\bot$ should be also in the set.   

\begin{definition}[Complete extension of defenses]
$D$ is complete iff $D$ is admissible,  and it satisfies the following conditions:  For all $z^\top\in   \dgn{\mathcal{F}}$,  $z^\top\in  D$.   For all $z^x_{\mathbf{y}}\in  \dgn{\mathcal{F}}$,  if $x\in \defendee{D}$,  and for all $z^{x^\prime}_{\mathbf{y^\prime}}\in  \dgn{\mathcal{F}}$,  $\mathbf{y^\prime} \not= \mathbf{y}$ implies $x^\prime\in \defendee{D}$,  then $z^x_{\mathbf{y}}\in D$.
\end{definition}

\begin{definition}[Preferred extension of defenses]
   $D$ is preferred iff $D$ is a maximal compete extension w.r.t.  set inclusion. 
\end{definition}

\begin{definition}[Grounded extension of defenses]
 
  $D$ is grounded iff $D$ is a minimal compete extension w.r.t.  set inclusion. 
 
\end{definition}

\begin{definition}[Stable extension of defenses]
 $D$ is stable iff $D$ is admissible,  and $\defendee{D}\cup \attacker{D} = \AR$. 
\end{definition}

 In this paper we use $\mathit{def}(\dgn{\mathcal{F}})$ to denote the set of extensions of defenses of ${\mathcal{F}}$ under semantics $\mathit{def}$,  where $\mathit{def}\in \{\mathit{CO},  \mathit{PR},  \mathit{GR},  \mathit{ST}\}$ denotes respectively complete,  preferred, grounded,  and stable defense semantics. 

\begin{example}
$\dgn{\mathcal{F}_{7}} =\{ a^{\top},  b^{\bot}_{\mathbf{a}},  c^{a}_{\mathbf{b}},  c^{\bot}_{\mathbf{g}},  d^{b}_{\mathbf{c}},  d^{g}_{\mathbf{c}},  e^{c}_{\mathbf{d}},  f^{d}_{\mathbf{e}},  g^{\top}\}$.  Then,  $\{a^{\top}, g^{\top},d^{g}_{\mathbf{c}},   f^{d}_{\mathbf{e}}\}$ is an extension under all semantics. 

  
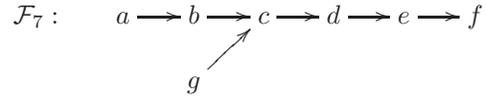
\begin{picture}(206,40)
\put(0,29){\xymatrix@C=0.55cm@R=0.38cm{
\mathcal{F}_{7}: &a\ar[r]&b\ar[r]&c\ar[r]&d\ar[r]&e\ar[r]&f
\\
 &&g\ar[ur]
 }}
\end{picture}
\end{example}

Note that there could be several complete extensions of defenses,  in which one may contain another.    

\begin{example}
About $\mathcal{F}_{8}$,  there are two complete extensions of defenses,  in which $ D_1 \subseteq D_2$:
\begin{itemize}
\item $D_1 = \{a^\top, d^\top,  c^a_{\mathbf{b}},  f^d_{\mathbf{e}}\}$
\item $ D_2 = \{a^\top, d^\top,  c^a_{\mathbf{b}},  c^d_{\mathbf{b}},  f^d_{\mathbf{e}}\}$
\end{itemize}
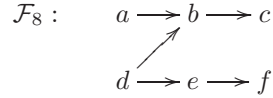
 \begin{picture}(206,38)
\put(0,29){\xymatrix@C=0.55cm@R=0.38cm{
\mathcal{F}_{8}: &a\ar[r]&b\ar[r]&c 
\\
 &d\ar[ur]\ar[r]&e\ar[r]&f
 }}
\end{picture}
\end{example}

Now, let us consider some properties of the defense semantics of an AF. 

The first property formulated in Theorem \ref{th-co2} is about the closure of defenses: If both $a^x_{\mathbf{z}}$ and $b^y_{\mathbf{z}^\prime}$ are in a complete extension of defenses, and $a^b_{\mathbf{z}^{\prime\prime}}$ is a defense,  then  $a^b_{\mathbf{z}^{\prime\prime}}$ is also in the same extension, where $x$ and $y$ are either $\top$ or some arguments in $\AR$.

\begin{theorem} \label{th-co2}
For all $D\in  \mathit{CO}(\dgn{\mathcal{F}})$,  $x,y\in\AR\cup\{\top\}$, if $a^x_{\mathbf{z}}$, $b^y_{\mathbf{z}^\prime}\in D$, $a^b_{\mathbf{z}^{\prime\prime}}\in \dgn{\mathcal{F}}$,  and $\mathbf{z}^\prime\neq \mathbf{z}^{\prime\prime}$,  then $a^b_{\mathbf{z}^{\prime\prime}}\in D$.
\end{theorem}

\begin{proof}
 Since $D$ is admissible,  according to the fourth item of the difintion of the admissible set of defenses,  it holds that $a^b_{\mathbf{z}^{\prime\prime}}\in D$.  
\end{proof}

\begin{example} \label{ex-closure}
$\dgn{\mathcal{F}_9} = \{a^\top,  b^\bot_{\mathbf{a}},   c^a_{\mathbf{b}},    d^b_{\mathbf{c}},  e^\top, f^\bot_{\mathbf{e}},  g^e_{\mathbf{f}},  g^c_{\mathbf{d}}\}$.  $  \mathit{CO}(\dgn{\mathcal{F}_9}) $ $= \{D\}$ where $D =\{ a^\top,     c^a_{\mathbf{b}},    e^\top,   g^e_{\mathbf{f}},  g^c_{\mathbf{d}}\}$.  Both $  c^a_{\mathbf{b}}$ and $g^e_{\mathbf{f}}$ are in $D$.  Since $ g^c_{\mathbf{d}}\in \dgn{\mathcal{F}_8}$, $ g^c_{\mathbf{d}}$ is in $D$.

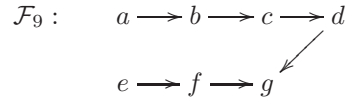
  \begin{picture}(206,38)
 \put(0,29){\xymatrix@C=0.55cm@R=0.38cm{
\mathcal{F}_9: &a\ar[r]&b\ar[r] & c \ar[r] & d \ar[dl]\\
& e\ar[r]&f\ar[r]&g
 }}
 \end{picture}
\end{example}

The second property formulated in Theorem \ref{th-1} is about the justfiability of defenses: If $z^x_{\mathbf{y}}$  is in a complete extension of defenses,  then there must be $x^u_{\mathbf{y}^\prime}$ in the same extension, where $u$ is either $\top$ or some argument in $ \AR$.

\begin{theorem}\label{th-1}
For all $D\in  \mathit{CO}(\dgn{\mathcal{F}})$,  if $z^x_{\mathbf{y}}\in D$ then $x^\top\in D$,  or there exists $u\in \AR$ such that $x^u_{\mathbf{y}^\prime}\in D$.
\end{theorem}

\begin{proof}
According to Definition \ref{def-ext-of-defences},  since $z^x_{\mathbf{y}}\in D$,  it holds that defender $x\in\defendee{D}$.  So,  there exists $u, y^\prime\in \AR$ such that $x^\top\in D$,  $x^u_{\mathbf{y}^\prime}\in D$,  or $x^\bot_{\mathbf{y}^\prime}\in D$.  Since $\bot\notin \defender{D}$,  $x^\bot_{\mathbf{y}^\prime}\notin D$.  As a result,  $x^\top\in D$,  or $x^u_{\mathbf{y}^\prime}\in D$.
\end{proof}

\begin{example} \label{ex-constr}
Consider  ${\mathcal{F}_{3}}$ again.  $ \mathit{CO}(\dgn{\mathcal{F}_{3}})= \{D_1, D_2, D_3\}$ where $D_1=\{ \}$, $D_2=\{a^{e}_{\mathbf{f}},   {d}^{a}_{\mathbf{c}},   {e}^{d}_{\mathbf{b}}\}$, $D_3=\{{c}^{f}_{\mathbf{a}},   {b}^{c}_{\mathbf{d}}, {f}^{b}_{\mathbf{e}}\}$.  
Take ${d}^{a}_{\mathbf{c}}$ in $D_2$
as an example: since $a$ is a defender in ${d}^{a}_{\mathbf{c}}$, there exists a defense whose defendee is $a$. In fact, $a^{e}_{\mathbf{f}}$ is in $D_2$. 
\end{example}

The third property formulated in Theorems \ref{prop-ad} and \ref{prop-adco} is about the relation between extensions of defenses and extensions of arguments, of an AF.  

\begin{theorem} \label{prop-ad}
For all $D \in \mathit{CO}(\dgn{\mathcal{F}})$,  $\defendee{D} \in \mathrm{co}(\mathcal{F})$.
\end{theorem}

\begin{proof}
Since $\defendee{D} \cap \attacker{D} = \emptyset$,  it holds that $\defendee{D}$ is conflict-free.  For all $z\in \defendee{D}$,  according to item $(ii)$,  $z$ is not attacked by an initial argument.  Then,  according to  item $(iii)$ and $(iv)$,  every attacker of $z$ is attacked by an argument in   $\defendee{D}$.  According to the definition of complete extension of defenses,  each argument that is defended by $\defendee{D}$ is in $\defendee{D}$.  
\end{proof}


\begin{theorem} \label{prop-adco}
For all $E \in \mathrm{co}(\mathcal{F})$, let $\mathrm{def}(E) = \{z^x_{\mathbf{y}}\mid z^x_{\mathbf{y}}\in \dgn{\mathcal{F}}: x,z\in E\} \cup \{x^\top \mid x^\top\in \dgn{\mathcal{F}}: x\in E\}$.  Then, $\mathrm{def}(E) \in  \mathit{CO}(\dgn{\mathcal{F}})$.
\end{theorem}

\begin{proof}
According to the definiton of complete extension of defenses,  this theorem holds.
\end{proof}

\begin{example}\label{ex-cr1}
Consider $\mathcal{F}_{10}$ 
below. We have:
\begin{itemize}
\item $\mathrm{co}(\mathcal{F}_{10}) = \{E_1, E_2\}$, where $E_1=\{ \}$, $E_2 =  \{b\}$;
\item   $\mathrm{def}(E_1) = \{ \}$, $\mathrm{def}(E_2) = \{ b^b_{\mathbf{a}}\}$;
\item $  \mathit{CO}(\dgn{\mathcal{F}_{10}}) = \{D_1, D_2\}$, where $D_1 = \{ \}$, $D_2 = \{b^b_{\mathbf{a}}\}$.
\end{itemize}

So, it holds that $\mathrm{def}(E_1) \in   \mathit{CO}(\dgn{\mathcal{F}_{10}}),  \mathrm{def}(E_2) \in   \mathit{CO}(\dgn{\mathcal{F}_{10}})$. 

  
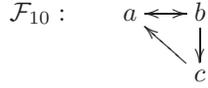
\begin{picture}(206,38)
\put(0,28){\xymatrix@C=0.55cm@R=0.38cm{
\mathcal{F}_{10}: &a\ar[r]&b\ar[l]\ar[d]
\\
 &&c\ar[ul]
 }}
\end{picture}
\end{example}

\section{Unsatisfiability and contraction of defenses}\label{ref-section-ucd}
In terms of defense semantics,  an interesting property is to exploit the local evaluation of defenses to simplify the computation of defense semantics,  based on the concept of unsatisfiability of some types of defenses.  In this paper,  as typical examples,  we introduce the types of defenses related to self-attacked arguments and arguments in 3-cycles. 

\begin{definition}[Unsatisfiability of defense]
We say that $z^x_{\mathbf{y}}$ is unsatisfiable iff $z^x_{\mathbf{y}}$ cannot be in any admissible set of defenses. 
\end{definition}

For an AF containing self-attacked arguments,  we have the following theorem.

\begin{theorem}
Defenses $z^\bot_{\mathbf{y}}$,  $z^y_{\mathbf{y}}$ and  $z^x_{\mathbf{z}}$are  unsatisfiable.   Furthermore,  if  $z^y_{\mathbf{y}}$ is a defense,  then $u^y_{\mathbf{v}}$ is  unsatisfiable.  
\end{theorem}

\begin{proof}
First,  obviously,  defense $z^\bot_{\mathbf{y}}$ is not satisfiable by definition. 

Second,  $z^y_{\mathbf{y}}$ means that $y$ self-attacks and it attacks $z$.  If $z^y_{\mathbf{y}}$ is in an admissible set of defenses,  then according to item $(iii)$ of the definition,  $y$ is a defendee.  According to item $(i)$,  $\defendee{D}\cap \attacker{D} \neq \emptyset$.  Contradiction.  

Third,  obviously,  $z^x_{\mathbf{z}}$ is not satisfiable. 

Fourth,  assume that $u^y_{\mathbf{v}}$ is in some admissible set $D$.  Then,  there exist some  $x, w\in AR$,  such that $y^x_{\mathbf{w}}$ is in $D$.  if $w = y$,  then this controdicts $\defendee{D}\cap \attacker{D} = \emptyset$.  Otherwise,  if $w \neq y$,  then $z^y_{\mathbf{y}}$ is also in $D$.  This also controdicts $\defendee{D}\cap \attacker{D} = \emptyset$.
\end{proof}

\begin{example}
$\dgn{\mathcal{F}_{2}} = \{{a}^{a}_{\mathbf{a}},  {b}^{a}_{\mathbf{a}},   {c}^{a}_{\mathbf{b}},  {d}^{b}_{\mathbf{c}}\}$. Then,  by removing all unsatisfiable defenses,  we get a contraction set of defenses   $\dgn{\mathcal{F}_{2}}^C = \{ {d}^{b}_{\mathbf{c}}\}$,  where $C = \{{a}^{a}_{\mathbf{a}},  {b}^{a}_{\mathbf{a}},   {c}^{a}_{\mathbf{b}}\}$.  The extensions based on these two sets are equivalent. 
\end{example}
 
Intuitively,  it is clear that this property will make the computation of defense extensions and also argument extensions much more efficient. 

Furthermore,  for a 3-cycle consisting of $x$,  $y$ and $z$ such that $x$ attacks $y$,  $y$ attacks $z$,  and $z$ attacks $x$,    we have the following theorem. 

\begin{theorem}
If there exist ${z}^{u}_{\mathbf{y}},  {y}^{z}_{\mathbf{x}}\in  \dgn{\mathcal{F}}$,  then  ${y}^{z}_{\mathbf{x}}$ is unsatisfiable. 
\end{theorem}

\begin{proof}
Assume that ${y}^{z}_{\mathbf{x}}$ is in some admissible set $D$.  Then,  for some $u$,  ${z}^{u}_{\mathbf{y}}$ is also in $D$.  As a result,  $\defendee{D} \cap \attacker{D} \supseteq \{y\}$,   contradicting $\defendee{D} \cap \attacker{D} = \emptyset$.
\end{proof}

Note that the unsatisfiability of a defense does not mean that its defendee is not accaptable.  See the following example.  In this case,  ${b}^{a}_{\mathbf{a}}$ is unacceptable,  but the argument $b$ is acceptable. 

 
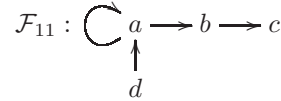
\begin{picture}(206,40)
\put(0,29){\xymatrix@C=0.55cm@R=0.38cm{
\mathcal{F}_{11}: &a\ar@(dl,ul)[]\ar[r]&b\ar[r]&c
\\
  &d\ar[u]
 }}
\end{picture}


When a set of defenses are unsatisfiable,  they can removed from the set of defenses,  resulting a contraction of the set of defenses.   

\begin{definition}[Contraction of set of defenses]
Let $\dgn{\mathcal{F}}$ the  set of defenses of $\mathcal{F}$, and $C\subseteq \mathcal{F}$ a set of defenses.  The contraction of $\dgn{\mathcal{F}}$ w.r.t. $C$,  denoted as $\dgn{\mathcal{F}}^C$,  is equal to $\dgn{\mathcal{F}}\setminus C$.
\end{definition}

\begin{theorem}
Let $\dgn{\mathcal{F}}$ the  set of defenses of $\mathcal{F}$,  and $\dgn{\mathcal{F}}^C$ the contraction of $\dgn{\mathcal{F}}$ w.r.t. $C$.  If for every defense in $C$,  it is unsatisfiable,  then $\mathit{def}(\dgn{\mathcal{F}}) = \mathit{def}( \dgn{\mathcal{F}}^C)$,  for all $\mathit{def}\in \{\mathit{CO},  \mathit{PR},  \mathit{GR}\}$. 
\end{theorem}

\begin{proof}
Obvious.
\end{proof}


\begin{example}
$\dgn{\mathcal{F}_{12}} = \{{d}^{\top},  {a}^{\bot}_{\mathbf{d}},  {a}^{b}_{\mathbf{c}},   {b}^{d}_{\mathbf{a}}, {b}^{c}_{\mathbf{a}}, {c}^{a}_{\mathbf{b}}\}$,  in which ${a}^{\bot}_{\mathbf{d}},  {a}^{b}_{\mathbf{c}},  {b}^{c}_{\mathbf{a}}$ and ${c}^{a}_{\mathbf{b}}$ are unacceptable.  So, the contraction of the set of defenses of $\mathcal{F}_{12}$ is $\dgn{\mathcal{F}_{12}}^C = \{{d}^{\top},      {b}^{d}_{\mathbf{a}}\}$, where $C = \{{a}^{\bot}_{\mathbf{d}},  {a}^{b}_{\mathbf{c}},   {b}^{c}_{\mathbf{a}}, {c}^{a}_{\mathbf{b}}\}$.  
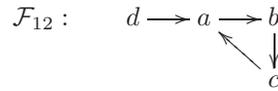
 \begin{picture}(206,40)
\put(0,29){\xymatrix@C=0.55cm@R=0.38cm{
\mathcal{F}_{12}: &d\ar[r]&a\ar[r]&b\ar[d]
\\
  &&&c\ar[lu]
 }}
\end{picture}

\end{example}

Note that this property is not affacted by the addition of some other defenses.  See the following example. 

 
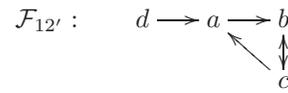
\begin{picture}(206,40)
\put(0,29){\xymatrix@C=0.55cm@R=0.38cm{
\mathcal{F}_{12^\prime}: &d\ar[r]&a\ar[r]&b\ar[d]
\\
  &&&c\ar[lu]\ar[u]
 }}
\end{picture}

In this case,  $\dgn{\mathcal{F}_{12^\prime}} = \{{d}^{\top},  {a}^{\bot}_{\mathbf{d}},  {a}^{b}_{\mathbf{c}},   {b}^{d}_{\mathbf{a}}, {b}^{c}_{\mathbf{a}}, {c}^{a}_{\mathbf{b}},  {b}^{b}_{\mathbf{c}}, {c}^{c}_{\mathbf{b}}\}$, and the contraction of the set of defenses of $\mathcal{F}_{12^\prime}$ is $\dgn{\mathcal{F}_{12^\prime}}^C = \{{d}^{\top},      {b}^{d}_{\mathbf{a}},   {b}^{b}_{\mathbf{c}}, {c}^{c}_{\mathbf{b}}\}$,  where $C =  \{{a}^{\bot}_{\mathbf{d}},     {a}^{b}_{\mathbf{c}},  {b}^{c}_{\mathbf{a}}, {c}^{a}_{\mathbf{b}}\}$.  


\section{Equivalence of AFs in terms of defenses}
The fourth property formulated in Theorems \ref{th-s-d} and \ref{th-s-d2} is about the equivalence of AFs under defense semantics, called \textit{defense equivalence of AFs}. 

\begin{definition}[Defense equivalence of AFs]
Let  $\mathcal{F}$  and $\mathcal{G}$ be two AFs.
 $\mathcal{F}$  and $\mathcal{G}$ are of defense equivalence w.r.t. a defense semantics $\mathit{def}$,  denoted as $\mathcal{F} \equiv^{\mathit{def}} \mathcal{G}$,  iff $\mathit{def}(\dgn{\mathcal{F}}) = \mathit{def}(\dgn{\mathcal{G}})$. 
\end{definition}

Concerning the relation between defense equivalence and standard equivalence of AFs, under complete semantics, we have the following theorem. 

\begin{theorem}\label{th-s-d}
Let  $\mathcal{F}$  and $\mathcal{G}$ be two AFs. If $\mathcal{F} \equiv^\mathit{CO} \mathcal{G}$, then $\mathcal{F} \equiv^\mathrm{co}  \mathcal{G}$.
\end{theorem}

\begin{proof}
If $\mathcal{F} \equiv^{\mathit{CO}} \mathcal{G}$, then $\mathit{CO}(\dgn{\mathcal{F}}) =\mathit{CO}(\dgn{\mathcal{G}})$.  Then,  it follows that $\mathrm{co}(\mathcal{F}) = \defendee{\mathit{CO}(\dgn{\mathcal{F}})} = \defendee{\mathit{CO}(\dgn{\mathcal{G}})} = \mathrm{co}(\mathcal{G})$.  Since  $\mathrm{co}(\mathcal{F}) =  \mathrm{co}(\mathcal{G})$, $\mathcal{F} \equiv^\mathrm{co}  \mathcal{G}$. $\square$
\end{proof}

Note that in many cases  $\mathcal{F} \equiv^\mathrm{co}  \mathcal{G}$,  but $\mathcal{F} \not\equiv^\mathit{CO} \mathcal{G}$. Consider the following example.

\begin{example} \label{ex-sdd}
Since $\mathrm{co}(\mathcal{F}_5) =  \mathrm{co}(\mathcal{F}_6) = \{\{a, c\}\}$, it holds that $\mathcal{F}_5 \equiv^\mathrm{co}  \mathcal{F}_6$.  Since $\mathit{CO}(\dgn{\mathcal{F}_5}) = \{a^\top,  {c}^{a}_{\mathbf{b}}\}$ and $\mathit{CO}(\dgn{\mathcal{F}_6}) = \{a^\top,  c^\top \}$,  $\mathit{CO}(\dgn{\mathcal{F}_5})\neq \mathit{CO}(\dgn{\mathcal{F}_6})$. So, it is not the case that $\mathcal{F}_5 \equiv^\mathit{CO} \mathcal{F}_6$.
\end{example}

About the relation between defense equivalence and strong equivalence of AFs, under complete semantics, we have the following lemma and theorem. 

\begin{lem} \label{lem-1}
It holds that $\mathit{CO}(\dgn{\mathcal{F}}) = \mathit{CO}(\dgn{\mathcal{F}^\mathrm{ck}})$.
\end{lem}

\begin{proof}
Since for every defense that is related to a self-attacked argument is unsatsifiable,  it is clear that $\mathit{CO}(\dgn{\mathcal{F}}) = \mathit{CO}(\dgn{\mathcal{F}^\mathrm{ck}})$.
\end{proof}

\begin{theorem}\label{th-s-d2}
Let  $\mathcal{F}$  and $\mathcal{G}$ be two AFs. If $\mathcal{F} \equiv_s^\mathrm{co} \mathcal{G}$, then $\mathcal{F} \equiv^\mathit{CO}  \mathcal{G}$.
\end{theorem}

\begin{proof}
Obvious.
\end{proof}

Note that in many cases  $\mathcal{F} \equiv^\mathit{CO}  \mathcal{G}$, but $\mathcal{F} \not\equiv_s^\mathrm{co} \mathcal{G}$.  Consider the following example.

\begin{example}
Since $\mathit{CO}(\dg{\mathcal{F}_{13}}) =\mathit{CO}(\dg{\mathcal{F}_{14}}) =  \{ \{a^\top$,  ${c}^{a}_{\mathbf{b}}\}  \}$,  $\mathcal{F}_{13} \equiv^\mathit{CO}  \mathcal{F}_{14}$.  However, since $\mathcal{F}_{13}^\mathrm{ck} \neq \mathcal{F}_{14}^\mathrm{ck}$, $\mathcal{F}_{13} \not\equiv_s^\mathrm{co}  \mathcal{F}_{14}$.

 
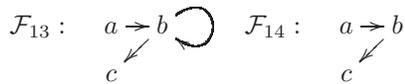
\begin{picture}(206,36)
 \put(0,24){\xymatrix@C=0.3cm@R=0.2cm{
\mathcal{F}_{13}:  & a\ar[r] &b\ar@(ur,dr)[] \ar[dl]&& \mathcal{F}_{14}: &a\ar[r] &b\ar[dl] \\
&c &&&& c
}}
 \end{picture}
\end{example}


Furthermore, 
defense semantics can be used to encode reasons for accepting arguments,  based on which  equivalence relation of root reasons can be defined.  Consider the following example.
\begin{example}\label{ex-eraa}
$\mathit{CO}(\dgn{\mathcal{F}_{15}}) = \{D_1, D_2\}$, where $D_1 = \{b^b_{\mathbf{a}}$, $d^b_{\mathbf{c}}$, $d^g_{\mathbf{c}},   g^e_{\mathbf{f}},  e^\top\}$, $D_2 = \{a^a_{\mathbf{b}}$, $d^g_{\mathbf{c}}$, $ g^e_{\mathbf{f}}$, $e^\top\}$.  One way to capture reasons for accepting arguments is to relate each reason to an extension of defenses.  For instance, concerning the reasons for accepting $d$ w.r.t. $D_1$,  we differentiate the following reasons:
\begin{itemize}
\item Direct reason: accepting \{b, g\} is a direct reason for accepting $d$. This reason can be identified in terms of defenses $d^b_{\mathbf{c}}$ and $d^g_{\mathbf{c}}$ in $D_1$. 
\item Root reason: accepting $\{e, b\}$ is a root reason for accepting $d$, in the sense that the elements of a root reason is either an initial argument, or an argument without further defenders except itself. This reason can be identified by means of viewing each defense as a binary relation in which only  defender and defendee in each defense are considered,    and allowing this relation to be transitive.  Given $\tuple{e,g}$ and $\tuple{g,d}$ according to $g^e_{\mathbf{f}}$ and $d^g_{\mathbf{c}}$ in $D_1$, we have $\tuple{e,d}$. Since $e$ is an initial argument, it is an element of the root reason. Given $d^b_{\mathbf{c}}$ in $D_1$, since $b$'s defender is  $b$ itself, $b$ is an element of the root reason. 
\end{itemize}

 
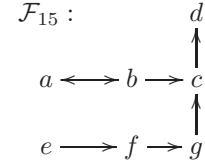
\begin{picture}(206,70) 
 \put(0,60){\xymatrix@C=0.45cm@R=0.4cm{
  \mathcal{F}_{15}:  &  & d 
\\
a\ar[r]&b\ar[l]\ar[r] &c \ar[u] 
\\
e\ar[r] &f \ar[r] & g\ar[u]
 \\
}} 
    \end{picture}
\end{example}

The informal notions in Example \ref{ex-eraa} are formulated as follows. 

\begin{definition}[Direct reasons for accepting arguments]\label{def-draa}
Let $\mathcal{F} = (\AR, \rightarrow)$ be an AF. Direct reasons for accepting arguments in $\mathcal{F}$ under a semantics $\mathit{def}$ is a function: 
\begin{equation}
\mathrm{dr}_\mathit{def}^{\mathcal{F}}: \AR \mapsto 2^{2^{\AR}} 
\end{equation}
For all $a\in \AR$, $\mathrm{dr}_\mathit{def}^{\mathcal{F}}(a) = \{\mathcal{DR}(a, D) \mid D\in \mathit{def}(\dgn{\mathcal{F}})\}$, where 
$\mathcal{DR}(a, D) = \{b\mid a^b_{\mathbf{c}}\in D\}$, if $a$ is not an initial argument; otherwise, $\mathcal{DR}(a, D) = \{\top\}.$
\end{definition}

\begin{example}
Continue Example \ref{ex-eraa}.  According to Definition \ref{def-draa},  under preferred semantics, $\mathrm{dr}^{\mathcal{F}_{11}}_\mathit{PR}(d) = \{R\}$, where $R = \{b,g\}$.  
\end{example}

\begin{definition}[Root reasons for accepting arguments] \label{def-rraa}
Let $\mathcal{F} = (\AR, \rightarrow)$ be an AF.  Root reasons for accepting arguments in $\mathcal{F}$ under a semantics $\mathit{def}$ is a function: 
\begin{equation}
\mathrm{rr}^{\mathcal{F}}_\mathit{def}: \AR \mapsto 2^{2^{\AR}} 
\end{equation}
For all $D\in\mathit{def}(\dgn{\mathcal{F}})$, we view $D$ as a transitive relation,  denoted as $\overline{D} = \{\tuple{x,z} \mid z^x_{\mathbf{y}}\in D \}$,  and let $\overline{D}^+$ be the transitive closure of $\overline{D}$. For all $\alpha\in \AR$, $\mathrm{rr}^{\mathcal{F}}_\mathit{def}(\alpha) = \{\mathcal{RR}(\alpha, D) \mid D\in \mathit{def}(\dgn{\mathcal{F}})\}$, where
\begin{eqnarray} \label{formula-rr}
\mathcal{RR}(\alpha, D)  =\{\alpha\mid \tuple{\alpha,\alpha}\in \overline{D}^+ \mbox{ or }  \tuple{\top,\alpha}\in \overline{D}^+\}  
\end{eqnarray}
if $\alpha$ is not an initial argument; otherwise, $\mathcal{RR}(\alpha, D) = \{\top\}$.
\end{definition}

\begin{example}
Continue Example \ref{ex-eraa}.  $\overline{D}_1^+ = \overline{D}_1 \cup \{\tuple{e,d}$, $\tuple{\top,g}$, $\tuple{\top,d}\}$; $\overline{D}_2^+ = \overline{D}_2 \cup \{\tuple{e,d}, \tuple{\top,g}, \tuple{\top,d}\}$. According to Definition \ref{def-rraa}, $\mathrm{rr}^{\mathcal{F}_{11}}_\mathit{PR}(d) = \{R\}$, where $R = \{b,e\}$.  
\end{example}

\begin{definition}[Root equivalence of AFs]
Let $\mathcal{F} = (\AR_1, \att_1)$ and $\mathcal{H} = (\AR_2, \att_2)$ be two AFs. For all $B\subseteq \AR_1\cap \AR_2$, if $B \neq \emptyset$, we say that $\mathcal{F}$ and $\mathcal{H}$ are equivalent w.r.t. the root reasons for accepting $B$ under semantics $\mathit{def}$, denoted $\mathcal{F}|B \equiv^\mathit{def}_{rr} \mathcal{H}|B$, iff for all $\alpha\in B$, $\mathrm{rr}_\sigma^{\mathcal{F}}(\alpha) = \mathrm{rr}_\sigma^{\mathcal{H}}(\alpha) $. 
\end{definition}

When $B = \AR_1 = \AR_2$, we write $\mathcal{F} \equiv^\mathit{def}_{rr} \mathcal{H}$ for $\mathcal{F}|B \equiv^\mathit{def}_{rr} \mathcal{H}|B$.

\begin{example}
Consider $\mathcal{F}_3$ and $\mathcal{F}_4$ in Section 1 again.  $\mathit{CO}(\dgn{\mathcal{F}_{3}})= \{D_1, D_2, D_3\}$ where $D_1=\{ \}$, $D_2=\{a^{e}_{\mathbf{f}},   {d}^{a}_{\mathbf{c}},   {e}^{d}_{\mathbf{b}}\}$, $D_3=\{{c}^{f}_{\mathbf{a}},   {b}^{c}_{\mathbf{d}}, {f}^{b}_{\mathbf{e}}\}$.  $\mathit{CO}(\dgn{\mathcal{F}_{4}})= \{D_4, D_5, D_6\}$ where $D_4=\{ \}$, $D_5=\{{a}^{a}_{\mathbf{b}}\}$,  $D_6=\{{b}^{b}_{\mathbf{a}}\}$.  Let $B = \{a, b\}$.
\begin{itemize}
\item $\mathrm{rr}_\mathit{CO}^{\mathcal{F}_1}(a) = \{\{ \}, \{a\}, \{ \}\}$,
\item $\mathrm{rr}_\mathit{CO}^{\mathcal{F}_1}(b) = \{\{\}, \{ \}, \{b\}\}$,
\item $\mathrm{rr}_\mathit{CO}^{\mathcal{F}_2}(a) =\{\{ \}, \{a\}, \{ \}\}$,
\item $\mathrm{rr}_\mathit{CO}^{\mathcal{F}_2}(b) = \{\{\}, \{ \}, \{b\}\}$.
\end{itemize}
So, it holds that $\mathcal{F}_3|B =^\mathit{CO}_\mathrm{rr} \mathcal{F}_4|B$.
\end{example}

\begin{theorem} \label{th-rr-st}
Let $\mathcal{F} = (\AR_1, \att_1)$ and $\mathcal{H} = (\AR_2, \att_2)$ be two AFs. If $\mathcal{F} \equiv^\mathit{CO}_{rr} \mathcal{H}$, then $\mathcal{F} \equiv^\mathrm{co}  \mathcal{H}$.
\end{theorem}

\begin{proof}
According to Definition \ref{def-rraa}, the number of extensions of $\mathrm{co}(\mathcal{F})$ is equal to the number of  $\mathrm{rr}_\mathit{CO}^{\mathcal{F}}(\alpha)$, where $\alpha\in \AR_1$. Since $\mathrm{rr}_\mathit{CO}^{\mathcal{F}}(\alpha) = \mathrm{rr}_\mathit{CO}^{\mathcal{H}}(\alpha)$, $\AR_1 = \AR_2$.

Let $\mathrm{rr}_\mathit{CO}^{\mathcal{F}}(\alpha) = \mathrm{rr}_\mathit{CO}^{\mathcal{H}}(\alpha) = \{R_1, \dots, R_n\}$. Let $\mathrm{co}(\mathcal{F}) = \{E_1, \dots, E_n\}$ be the set of extensions of $\mathcal{F}$, where $n \ge 1$. 

For all $\alpha\in \AR_1$, for all $R_i$, $i = 1, \dots, n$, we have $\alpha\in E_i$ iff  $R_i \neq \{\}$, in that in terms of Definition \ref{def-rraa},when $R_i \neq \{\}$, there is a reason to accept $\alpha$. 

On the other hand, let $\mathrm{co}(\mathcal{H}) =  \{S_1, \dots, S_n\}$ be the set of extensions of $\mathcal{H}$. For all $\alpha\in \AR_2 =\AR_1$, for all $R_i$, $i = 1, \dots, n$, for the same reason, we have $\alpha\in S_i$ iff  $R_i \neq \{\}$. So, it holds that $E_i = S_i$ for $i = 1, \dots, n$, and hence $\mathrm{co}(\mathcal{F}) = \mathrm{co}(\mathcal{H})$, i.e.,  $\mathcal{F} \equiv^\mathrm{co}  \mathcal{H}$. $\square$
\end{proof}

Note that in many cases  $\mathcal{F} \equiv^\mathrm{co}  \mathcal{H}$, but $\mathcal{F} \not\equiv^\mathit{CO}_{rr} \mathcal{H}$. This can be easily verified by considering $\mathcal{F}_5$ and $\mathcal{F}_6$ in Example \ref{ex-sdd}.

The notion of root equivalence of AFs can be used to capture a kind of  summarization in the graphs. Consider the following example borrowed from \cite{iomultipoles}.

\begin{example} \label{ex-sum1}
Let $\mathcal{F}_{16} = (\AR, \rightarrow)$ and $\mathcal{F}_{17} =(\AR^\prime, \rightarrow^\prime)$, illustrated below. Under complete semantics, $\mathcal{F}_{17}$ is a summarization of $\mathcal{F}_{16}$ in the sense that $\AR^\prime \subseteq \AR$, and the root reason of each argument in $\mathcal{F}_{17}$ is the same as that of each corresponding argument in $\mathcal{F}_{16}$. More specifically, it holds that $\mathrm{rr}^{\mathcal{F}_{16}}_\mathrm{co}(e_3) = \mathrm{rr}^{\mathcal{F}_{17}}_\mathrm{co}(e_3) = \{\{e_1, e_2\}\}$, $\mathrm{rr}^{\mathcal{F}_{16}}_\mathrm{co}(e_2) = \mathrm{rr}^{\mathcal{F}_{17}}_\mathrm{co}(e_2) = \{\{\top\}\}$, and $\mathrm{rr}^{\mathcal{F}_{16}}_\mathrm{co}(e_1) = \mathrm{rr}^{\mathcal{F}_{17}}_\mathrm{co}(e_1) = \{\{\top\}\}$.

 \begin{picture}(206,36)
 \put(0,24){\xymatrix@C=0.26cm@R=0.2cm{
\mathcal{F}_{16}:  & e_1\ar[r]&a_1\ar[r]&a_2 \ar[r]&o\ar[r]& e_3  & \mathcal{F}_{17}: & e_1\ar[r]&o\ar[r]&e_3 \\
&e_2\ar[r]&b_1\ar[r]&b_2\ar[ru]&  &&& e_2\ar[ru]
}}
 \end{picture}
\end{example}

Formally, we have the following definition. 

\begin{definition}[Summarization of AFs]
Let $\mathcal{F} = (\AR_1, \att_1)$ and $\mathcal{H} = (\AR_2, \att_2)$ be two AFs. $\mathcal{F}$ is a summarization of $\mathcal{H}$ under a semantics $\mathit{def}$ iff $\AR_1 \subset \AR_2$, and $\mathcal{F}|\AR_1 \equiv^\mathit{def}_{rr} \mathcal{H}|\AR_1$.
\end{definition}




\section{Conclusions}
In this paper, we have proposed a defense semantics of argumentation based on a novel notion of defense,  and used it to study contraction of defenses and equvalence relations between AFs.  By introducing two new kinds of equivalence relation between AFs, i.e., defense equivalence and root equivalence, we have shown that defense semantics can be used to capture the equivalence of AFs from the perspective of reasons for accepting arguments. In addition, we have defined a notion of summarization of AFs by exploiting root equivalence.  

Since defense semantics explicitly represents defense relation in extensions and can be used to encoded reasons for accepting arguments, it provides a new way to investigate such topics as summarization in argumentation, dynamics of argumentation, dialogical argumentation \cite{DBLP:journals/ijar/HunterT16,DBLP:conf/atal/FanT16}, etc.  Further work on these topics is promising. In addition, it might be interesting to study defense semantics beyond Dung's argumentation, including ADFs \cite{DBLP:conf/kr/BrewkaW10}, bipolar argumentation \cite{DBLP:journals/ijar/CayrolL13}, structured argumentation \cite{DBLP:journals/argcom/BesnardGHMPST14}, etc.

\bibliographystyle{named}

\begin{thebibliography}{99}

\bibitem{Baroni:KER}
Pietro Baroni, Martin Caminada, and Massimiliano Giacomin. 
\newblock An introduction to argumentation semantics. 
\newblock The Knowledge Engineering Review, 26(4):365–410, 2011.

\bibitem{iomultipoles}
Pietro Baroni, Guido Boella, Federico Cerutti, Massimiliano Giacomin, Leendert van der Torre, and Serena Villata. 
\newblock On the input/output behavior of argumentation frameworks. 
\newblock Artificial Intelligence, 217:144– 197, 2014.

\bibitem{DBLP:journals/ai/BaumannBU22}
Ringo Baumann, Gerhard Brewka, and Markus Ulbricht. 
\newblock  Shedding new light on the foundations of abstract argumentation: Modularization and weak admissibility. 
\newblock Artif. Intell., 310:103742, 2022.

\bibitem{DBLP:journals/argcom/BesnardGHMPST14}
 Philippe Besnard, Alejandro Javier Garc\'{i}a, Anthony Hunter, Sanjay Modgil, Henry Prakken, Guillermo Ricardo Simari, and Francesca Toni. \newblock  Introduction to structured argumentation. 
\newblock Argument \& Computa- tion, 5(1):1–4, 2014.

\bibitem{DBLP:conf/kr/BrewkaW10}
 Gerhard Brewka and Stefan Woltran. 
\newblock Abstract dialectical frameworks. 
\newblock In Principles of Knowledge Representation and Reasoning: Proceedings of the Twelfth International Conference, KR 2010, Toronto, Ontario, Canada, May 9-13, 2010, 2010.

\bibitem{DBLP:journals/ijar/CayrolL13}
Cayrol and Lagasquie-Schiex, 2013 Claudette Cayrol and Marie-Christine Lagasquie-Schiex. 
\newblock Bipolarity in argumentation graphs: Towards a better understanding. 
\newblock Int. J. Approx. Reasoning, 54(7):876–899, 2013.

\bibitem{DBLP:journals/ai/CharwatDGWW15}
Gu{n}ther Charwat, Wolfgang Dvorak, Sarah Alice Gaggl, Johannes Peter Wallner, and Stefan Woltran. 
\newblock Methods for solving reasoning problems in abstract argumentation - A survey. 
\newblock Artif. Intell., 220:28–63, 2015.

\bibitem{DBLP:journals/ai/Dung95}
Phan Minh Dung. 
\newblock On the acceptability of arguments and its fundamental role in nonmonotonic reasoning, logic programming and n-person games. 
\newblock Artificial Intelligence, 77(2):321–358, 1995.

\bibitem{DBLP:conf/atal/FanT16}
Xiuyi Fan and Francesca Toni. 
\newblock On the interplay between games, argumentation and dialogues. 
\newblock In Proceedings of the 2016 International Conference on Au- tonomous Agents \& Multiagent Systems, Singapore, May 9-13, 2016, pages 260–268, 2016.

\bibitem{DBLP:journals/ijar/HunterT16}
Anthony Hunter and Matthias Thimm. 
\newblock Optimization of dialectical outcomes in dialogical argumentation. 
\newblock Int. J. Approx. Reasoning, 78:73–102, 2016.

\bibitem{DBLP:journals/amai/Liao13}
 Beishui Liao. 
\newblock Toward incremental computation of argumentation semantics: A decomposition-based approach. Ann. Math. 
\newblock Artif. Intell., 67(3-4):319–358, 2013.

\bibitem{DBLP:conf/kr/OikarinenW10}
Emilia Oikarinen and Stefan Woltran. 
\newblock Characterizing strong equivalence for argumentation frameworks. In Principles of Knowledge Representation and Reasoning:
\newblock  Proceedings of the Twelfth International Conference, KR 2010, Toronto, Ontario, Canada, May 9-13, 2010, 2010.

\end{thebibliography}

\end{document}